# Reversible data hiding with dual pixel-value-ordering and minimum prediction error expansion

Md. Abdul Wahed[1,2, ‡*] Hussain Nyeem[1,2,‡],

**1** Department of Electrical, Electronic and Communication Engineering (EECE)
Military Institute of Science and Technology (MIST)
Mirpur Cantonment, Dhaka-1216, Bangladesh
**2** Department of EECE, Engineering Faculty
Bangladesh Military Academy (BMA), Chattogram-4315, Bangladesh

‡These authors contributed equally to this work.
*wahedruet@gmail.com

## Abstract

Pixel Value Ordering (PVO) holds an impressive property for high fidelity Reversible Data Hiding (RDH). In this paper, we introduce a dual-PVO (dPVO) for Prediction Error Expansion (PEE), and thereby develop a new RDH scheme to offer a better rate-distortion performance. Particularly, we propose to embed in two phases: *forward* and *backward*. In the *forward* phase, PVO with classic PEE is applied to every non-overlapping image-block of size $1 \times 3$. In the *backward* phase, *minimum-set* and *maximum-set* of pixels are determined from the pixels predicted in the forward phase. The *minimum* set only contains the lowest predicted pixels and the *maximum* set contains the largest predicted pixels of each image-block. Proposed dPVO with PEE is then applied to both sets, so that the pixel-values of *minimum* set are increased and that of the *maximum* set are decreased by a unit value. Thereby, the pixels predicted in the *forward* embedding can partially be restored to their original values resulting in both a better-embedded image quality and a higher embedding rate. Experimental results have recorded a promising rate-distortion performance of our scheme with a significant improvement of embedded image quality at higher embedding rates compared to the popular and state-of-the-art PVO-based RDH schemes.

## Introduction

Reversible Data Hiding (RDH) is an evolving covert-communication technology [1]. It can imperceptibly embed *data* in a given *cover* media to output an *embedded* media, and losslessly retrieve both the embedded data and cover image later on demand basis [2]. An RDH application framework aims at achieving different security properties like authentication and/or integrity verification of the multimedia information [3]. Those properties can be readily attained by using standard cryptographic techniques. However, such direct application of cryptographic techniques obliterates the semantic understanding of the media due to the (temporary) loss of the spatial information. An RDH scheme helps avoid this obliteration problem as such it can borrow the security properties of the cryptographic techniques by embedding a 'secret' data imperceptibly. Here, a pre-processing of the data requires to employ a suitable cryptographic technique for a data hiding application, which is beyond the scope of this paper. Additionally, we also restrict our attention in this paper to the digital image application of an RDH scheme, which may principally be extended to different digital media (*e.g*., image, video, audio or speech).

Development of an RDH scheme is generally steered by a higher embedding capacity with the invertible and minimum-possible distortion [4]. For example, the pioneering difference expansion (DE)-based RDH scheme [5] with invertible distortion was immediately improved for higher capacity with generalized expansion [6], reduced location map [7, 8], sorting and prediction [9, 10], and adaptive embedding [11, 12]. On the other hand, for minimizing distortion, the histogram shifting (HS)-based scheme [13] was improved using the difference-histogram [14–16] and multiple histograms [17, 18]. Other potential developments include the RDH schemes with prediction error expansion (PEE) [19–39], vector quantization [40, 41], interpolation [42–44], encryption [45, 46], and transform techniques [47, 48].

Among the varieties of RDH principle, PEE is much investigated for its efficient rate-distortion performance [49, 50]. It can better utilize the combined principles of DE and HS to expand prediction errors for data hiding. Unlike the use of pixel-histogram in basic HS, it deals with the prediction errors to obtain a much sharper histogram with a set of higher peak beans resulting in higher embedding capacity. Additionally, unlike the direct change of pixels in basic DE, it expands the prediction errors to offer minimum possible changes in the pixels resulting in higher quality embedded images. Further developments of the PEE-based schemes can also be tracked with the context modification [20, 21, 51], prediction error classification [22–24], adaptive image-block size [25, 26], two-dimensional histogram modification [28], pair-wise PEE [27, 35] and pixel value ordering [19, 29–34, 36–38, 52–54].

Pixel value ordering (PVO) has been a prominent solution to minimize the prediction errors in PEE. Although the pixel value selection, grouping and sorting principles for prediction were utilized in [9, 19], the principle of PVO was well established by Li *et al*. [29]. That scheme predicted the maximum-minimum pixel pairs to embed with lower distortion. Peng *et al*. [30] improved the PVO with a new histogram-modification principle. Ou *et al*. [31] extended the basic PVO to PVO-*k* for adaptive embedding in the blocks according to the numbers of maximum- and minimum-valued pixels. Unlike the block-wise prediction in the original PVO, Qu *et al*. [32] then extended it to be a pixel-wise for a larger capacity and better image fidelity. Wang *et al*. [33] introduced a dynamic partitioning to construct image-blocks for PVO according to the blocks' complexity to improve the embedding capacity. Other recent developments include multiple histograms modification [55], pair-wise PEE [34, 56] and multi-pass adaptive PVO [38, 53, 54].

The PVO-based RDH schemes mentioned above demonstrated a better rate-distortion performance for lower embedding capacity requirement. However, their rate-distortion performances sharply decrease with higher embedding rate. Their maximum embedding capacity limits are also lower and they mostly rely on the complex and recursive embedding conditions. Thus, an additional embedding level with counter-balancing of the distortion caused by the expansion of prediction errors have been introduced in [52]. In this paper, we further investigate the counter-balancing approach with substantial statistical analysis in developing dual-PVO (dPVO) embedding for both the better image quality and higher embedding rate is introduced.

The main contribution of the proposed dPVO scheme is summarized as follows.

- Unlike the existing PVO-based schemes, the dPVO is developed to embed in two phases: (*i*) forward embedding, and (*ii*) backward embedding with minimal pixel grouping. While the forward embedding applies PEE with PVO to every non-overlapping image-block of size $1 \times 3$, the backward embedding partitions the previously predicted pixels into *minimum-set* and *maximum-set* for embedding using our new embedding technique called dual-PVO (dPVO) with pairwise PEE.

- The proposed *backward* embedding with dPVO and PEE is designed to counterbalance the distortion caused in the *forward* embedding phase. In other words, The pixels predicted in the first phase thus can be partially restored to their original values. As a result, unlike the conventional PEE-based RDH schemes (where embedding capacity costs image quality), embedding capacity or rate is increased (*i.e.*, 11.98% on average) in the backward embedding at a better embedded image quality (*i.e.*, 2.16% on average) (see Table 1 in Section **Experimental results and analysis**).

- The experimental results also demonstrate a significantly better image fidelity at the higher embedding rate (*i.e.*, above 10,000 bits), while it can also maintains competitive embedding rate-distortion performance at the lower embedding capacity (*i.e.*, up to 10,000 bits) over the popular and state-of-the-art PVO-based RDH schemes [19, 29, 30, 32–34, 36, 38] (see Section **Experimental results and analysis**).

The remainder of this paper is structured as follows. Related RDH schemes are briefly reviewed in Section **Related PVO-based RDH schemes**. Our new RDH scheme is presented with required computational details in Section **A new PVO-based RDH scheme**. Rate-distortion performance of our scheme for different benchmark image-sets is evaluated, analyzed and validated in Section **Experimental results and analysis**. Conclusions are given in Section **Conclusions**.

## Related PVO-based RDH schemes

We now briefly introduce the basic principle of PVO [29] and its successive development to the Jung's minimum-block PVO based RDH scheme [36]. In what follows, we present a generalized framework of PVO-based embedding that is conventionally used to define the principle of PVO and its improvements in existing schemes [19, 29, 30, 32–34, 36, 38].

A PVO-based embedding generally starts with partitioning a cover image $I$ of size $M \times N$ into a set of non-overlapping image-blocks, *i.e.*, $I = [X_k]$. With each image-block containing $n$ pixels, *i.e.*, $X = (x_1, x_2, \ldots x_n)$, total number of image-blocks is $k = \frac{M \times N}{n}$. For each image-block $X$, its pixels, $(x_1, x_2, \cdots x_n)$ are now sorted in ascending order using a sorting function $\sigma(\cdot)$ to output $(x_{\sigma(1)}, x_{\sigma(2)}, \cdots x_{\sigma(n)})$. The function, $\sigma : \{1, 2, \cdots n\} \rightarrow \{1, 2, \cdots n\}$ is a unique one-to-one mapping such that $x_{\sigma(1)} \leq x_{\sigma(2)} \leq \cdots \leq x_{\sigma(n)}$ with $\sigma(i) < \sigma(j)$ if $x_{\sigma(i)} = x_{\sigma(j)}$ and $i < j$. Once the image-block pixels are sorted, they are used for prediction with a suitable PEE-based embedding of *data* leading to the development of different PVO-based schemes as follows. Without loss of generality, we illustrate different PEE-based embedding conditions below for a data bit, $b \in \{0, 1\}$.

### Li et al.'s PVO-based RDH scheme

Li *et al*. [29] proposed to use the second maximum block-pixel, *i.e.*, $x_{\sigma(n-1)}$ to predict the maximum $x_{\sigma(n)}$. The prediction error, $e$ is computed using Eq. (1). PEE-based embedding is then carried out with the histogram of $e$. Only bin 1 with $e = 1$ is used for embedding of a data-bit, $b$, and other bins are expanded with the condition of higher values of $e$ in Eq. (2a). With the modified error, $\hat{e}$, the maximum block-pixel is updated accordingly as in Eq (2c).

$$e = x_{\sigma(n)} - x_{\sigma(n-1)} \tag{1}$$

$$\hat{e} = \begin{cases} e, & \text{if } e = 0 \\ e + b, & \text{if } e = 1 \\ e + 1, & \text{if } e > 1 \end{cases} \quad (2a)$$

$$\hat{x}_{\sigma(n)} = x_{\sigma(n-1)} + \hat{e} \quad (2b)$$

$$= \begin{cases} x_{\sigma(n)}, & \text{if } e = 0 \\ x_{\sigma(n)} + b, & \text{if } e = 1 \\ x_{\sigma(n)} + 1, & \text{if } e > 1 \end{cases} \quad (2c)$$

This process of PEE embedding repeats for all the blocks (until the last data-bit is embedded assuming the embedding capacity requirement is attained) to output an embedded image, $\hat{I}$. It is apparent that this embedding does not change the pixel value order after embedding to ensure perfect data extraction and lossless recovery.

Embedded data are extracted with the inverse embedding conditions. Specifically, with the embedded image, $\hat{I}$, image-blocks $\hat{X}_k$ are obtained. For each block, $\hat{X}$, the prediction errors are re-generated using Eq. (3a). The extracted data-bits and the original pixels are obtained using the reverse PEE conditions in Eq. (3).

$$\hat{e} = \hat{x}_{\sigma(n)} - \hat{x}_{\sigma(n-1)} \quad (3a)$$

$$\text{if } \hat{e} \in \{1, 2\} : \begin{cases} b = \hat{e} - 1 \\ x_{\sigma(n)} = \hat{x}_{\sigma(n)} - b \end{cases} \quad (3b)$$

$$\text{if } \hat{e} > 2 : x_{\sigma(n)} = \hat{x}_{\sigma(n)} - 1 \quad (3c)$$

$$\text{if } \hat{e} = 0 : x_{\sigma(n)} = \hat{x}_{\sigma(n)} \quad (3d)$$

With the maximum possible change to a pixel value by 1, the embedded image quality also remains high. With the consideration of minimum block-pixels, this basic PVO-based embedding is further improved in [30, 36].

### Jung's minimum PVO-based RDH scheme

Jung [36] recently proposed a minimal case of the PVO-based RDH scheme to embed 2 bits in an image-block of size $1 \times 3$. Particularly, an image $I$ is partitioned into a set of image-blocks containing three pixels as such $I = [X_k]$ with $k \in \{1, 2, \ldots, \frac{M \times N}{3}\}$. This means that, with the general PVO framework presented at the beginning of this section, Jung's scheme operates on each image-block $X$ with the number of block-pixels, $n = 3$. Thus the sorting function, $\sigma()$ is used to sort the block-pixels $(x_1, x_2, x_3)$ to be $(x_{\sigma(1)}, x_{\sigma(2)}, x_{\sigma(3)})$, where $x_{\sigma(3)}$ and $x_{\sigma(1)}$ are the maximum and minimum block-pixels, respectively. A pair of prediction errors, $e_{max}$ and $e_{min}$ for each image-block is calculated from the middle block-pixel, $x_{\sigma(2)}$ according to the Eq. (4a) and Eq. (4b). These errors are expanded with the embedding of a data-bit, $b$ or shifting by the value 1 with Eq. (4c) and Eq. (4d). The maximum and minimum block-pixels are then predicted from the middle block-pixel and the expanded errors with Eq. (4e) and Eq. (4f), respectively.

$$e_{max} = x_{\sigma(3)} - x_{\sigma(2)} \tag{4a}$$

$$e_{min} = x_{\sigma(1)} - x_{\sigma(2)} \tag{4b}$$

$$\hat{e}_{max} = \begin{cases} e_{max}, & \text{if } e_{max} = 0 \\ e_{max} + b, & \text{if } e_{max} = 1 \\ e_{max} + 1, & \text{if } e_{max} > 1 \end{cases} \tag{4c}$$

$$\hat{e}_{min} = \begin{cases} e_{min}, & \text{if } e_{min} = 0 \\ e_{min} - b, & \text{if } e_{min} = -1 \\ e_{min} - 1, & \text{if } e_{min} < -1 \end{cases} \tag{4d}$$

$$\hat{x}_{\sigma(3)} = x_{\sigma(2)} + \hat{e}_{max} \tag{4e}$$

$$\hat{x}_{\sigma(1)} = x_{\sigma(2)} + \hat{e}_{min} \tag{4f}$$

The data extraction and original block-pixels' recovery follow the inverse PEE embedding principle of the Jung's scheme in Eq. (5) like other PVO-based RDH schemes. With the recovery of the maximum and minimum block-pixels of all expanded pixels, the original image is recovered. At the same time, the data-bits are extracted from each embedded image-blocks and concatenated to get the original data.

$$b = \begin{cases} \hat{e}_{max} - 1, & \text{if } 1 \leq \hat{e}_{max} \leq 2 \\ -\hat{e}_{min} - 1, & \text{if } -2 \leq \hat{e}_{min} \leq -1 \end{cases} \tag{5a}$$

$$x_{\sigma(3)} = \begin{cases} \hat{x}_{\sigma(3)}, & \text{if } \hat{e}_{max} = 0 \\ \hat{x}_{\sigma(3)} - b, & \text{if } 1 \leq \hat{e}_{max} \leq 2 \\ \hat{x}_{\sigma(3)} - 1, & \text{if } \hat{e}_{max} > 2 \end{cases} \tag{5b}$$

$$x_{\sigma(2)} = \hat{x}_{\sigma(2)} \tag{5c}$$

$$x_{\sigma(1)} = \begin{cases} \hat{x}_{\sigma(1)}, & \text{if } \hat{e}_{min} = 0 \\ \hat{x}_{\sigma(1)} + b, & \text{if } -2 \leq \hat{e}_{min} \leq -1 \\ \hat{x}_{\sigma(1)} + 1, & \text{if } \hat{e}_{min} < -2 \end{cases} \tag{5d}$$

With a single reference pixel in an image-block, Jung's scheme predicts the maximum and minimum block-pixels as such in every three pixels of an image, two bits of data can be embedded. Thus, the embedding capacity is improved with a reasonably good embedded image quality. However, the overall embedding rate-distortion performance at lower embedding rate is still much lower than the advanced PVO-based RDH schemes [32, 34, 38]. In this paper, more effective use of the Jung's PVO is investigated and thus the development of a higher capacity RDH scheme with a competitive visual quality of an embedded image is presented in the section below.

## A new PVO-based RDH scheme

In this section, we introduce a new RDH scheme with *dual pixel value ordering* (dPVO) and PEE. A PVO-based embedding has evolved to utilize image correlations for a better possible rate-distortion performance, as mentioned in Sec. **Introduction**. With classic PVO, pixel values in an image-block are kept unchanged or expanded (either for embedding or shifting) centering the reference pixel(s). This principle of embedding has been better utilized with the adaptive size of image-block or multilevel embedding in the recent schemes for a better rate-distortion performance. However, expanded pixels have not been considered yet for reverse expansion to restore them to their respective original pixel values partially. In this paper, we attempt to utilize this reverse expansion property in the 'backward embedding' with a minimum PVO scenario of the Jung's scheme [36]. Thereby, we show that such reverse expansion of the expanded pixels can further improve the rate-distortion performance.

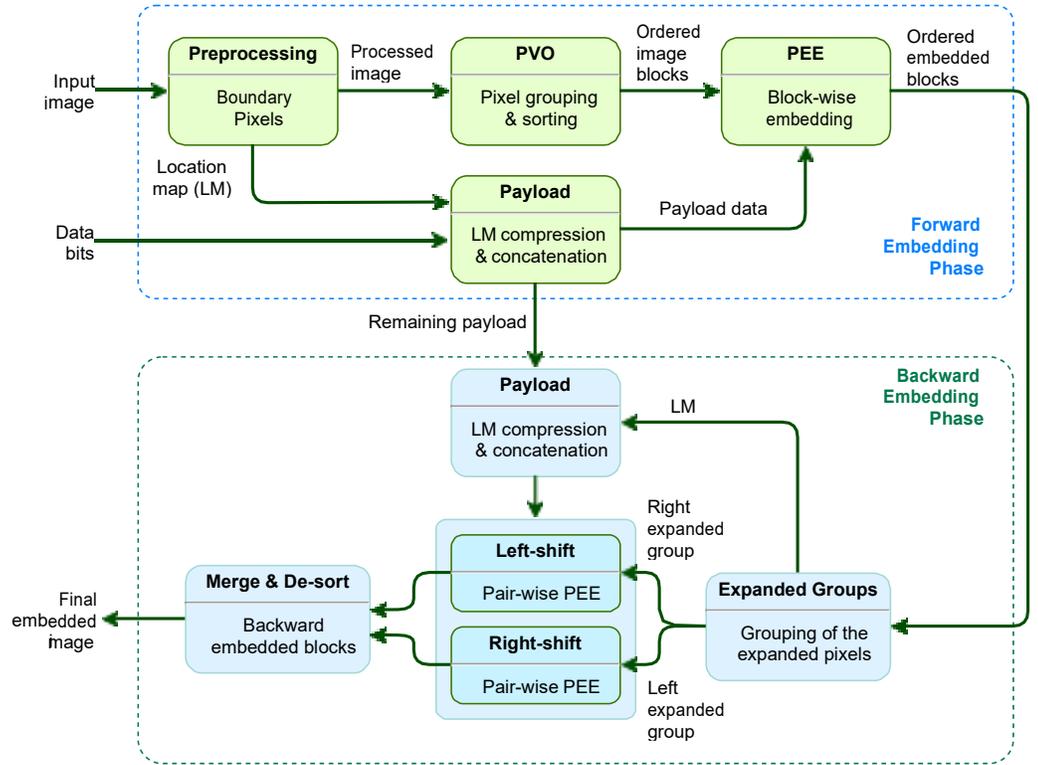

**Fig 1. A general framework of the proposed dPVO-based embedding**

Our RDH scheme constitutes two phases of embedding; namely, (*i*) forward embedding with PVO and PEE, and (*ii*) backward embedding with dPVO and pairwise-PEE. These two phases of embedding are expected to improve both the visual quality of the embedded image and the embedding rate. Extraction of our scheme, on the other hand, follows the inverse processing of those two-phases of embedding. A high-level conceptual model of these phases of embedding is presented in Fig 1. The input image is first pre-processed modifying the boundary pixels with intensity values of 255 or 0 to tackle any overflow/ underflow situation. A location map is generated to keep track of the boundary pixels, which is compressed and appended with the data-bits.

The pre-processed image undergoes forward embedding yielding sorted embedding blocks. In the backward phase, block-wise pixel grouping is done for obtaining a minimum group and a maximum group of embedded pixels. A location map is generated for tracking the skipped pixels and its compressed version is appended with data. The minimum and maximum groups obtained then undergo pair-wise PEE. Finally, merging and de-sorting yields the final embedded image. Additionally, the auxiliary information required for extracting the compressed location map is embedded in the first LSBs of the border pixels. The original LSBs are recorded in a binary sequence and appended with data. For data extraction and image recovery, an exact reverse process is followed. Rest of this section explains the embedding and extraction processes with more computational details followed by a working example of the proposed scheme.

## Forward embedding

As mentioned above, for forward embedding with PEE, we employ the Jung's PVO-based scheme that starts with partitioning an input image, $I$ into a set of non-overlapping image-blocks of size $1 \times 3$. This is discussed in Sec. **Jung's minimum PVO-based RDH scheme**. Each block-pixels $(x_1, x_2, x_3)$ are sorted to obtain $(x_{\sigma(1)}, x_{\sigma(2)}, x_{\sigma(3)})$, where $x_{\sigma(1)}$ and $x_{\sigma(3)}$ are the minimum and maximum block-pixels, respectively. With the computation and expansion of the pair of prediction errors, $e_{min}$ and $e_{max}$ using Eq. (4a) to Eq. (4d), either a data-bit, $b$ is embedded or error-value is shifted by 1. The minimum and maximum block-pixels are then predicted from the middle block-pixel and the expanded errors with Eq. (4e) and Eq. (4f).

We note that the overflow and underflow problem is usually tackled with the conventional process of recording a location map $M_k$ for the $k$-th image-block (for example, see Ref. [37,38]). The map is initialized as an empty-set for each $k$-th image-block, and for each pixel of the block, we append either a '1' for a boundary pixel or '0' for any other pixel to $M_k$, Continuing this for all $k$, a complete location map $M_{ou} = \{M_k\}$ is obtained, losslessly compressed using arithmetic coding and appended to the embedding data-bits. With an input image of bit-depth 8-bit, for example, a boundary pixel, $x$ in an image-block is then updated using Eq. (6).

$$x = \begin{cases} x - 1, & \text{if } x = 255 \\ x + 1, & \text{if } x = 0 \\ x, & \text{otherwise} \end{cases} \quad (6)$$

Given the input image, $I$ and a set of data-bits, $D_f$, with this forward embedding, we thus obtain the embedded image, $\hat{I}$. For simplicity, we omit the notational difference between the original input image and its pre-processed version with modified boundary pixels.

## Backward embedding

This embedding operates on $\hat{I}$ and aims to restore the changes (*i.e.*, expansion) made in the forward embedding. It is obvious that the maximum and minimum pixels of an image-block, *i.e.* $x_{\sigma(1)}$ and $x_{\sigma(3)}$, can experience a maximum expansion of value 1 either for embedding of a data-bit '1' or for shifting the pixel by the value 1. Thus, all the predicted pixels that experience this expansion are considered for minimum and maximum groups, $\hat{X}_{min}$ and $\hat{X}_{max}$, respectively. Additionally, we locate the predicted pixels that remain unchanged in the first phase of embedding by recording their location map in $LM$. Computing of $\hat{X}_{min}$ and $\hat{X}_{max}$ is presented in Algorithm 1.

Particularly, our proposed dPVO first separates two sets of pixels expanded in the forward embedding. With a classic PVO on an image-block of size $1 \times 3$, the lowest and highest pixels remain in the same order after their left-and right-ward expansion. So, the lowest and highest predicted pixels of all image-blocks can be separated into two sets as such applying a pairwise PVO based backward embedding on these two sets can restore their original pixel values. For example, in the forward embedding, an embedded image, $\hat{I}$ is obtained with the Jung's scheme. In the backward embedding with dPVO, a set $\hat{X}_{min}$ is computed with the lowest predicted pixels, $\{x_{\sigma(1)}\}$ of all the blocks $[\hat{X}_k]$. Similarly, another set $\hat{X}_{max}$ captures the largest predicted pixels, $\{x_{\sigma(3)}\}$ of all the blocks $[\hat{X}_k]$. All pairs of pixel values in each of these two sets, $\hat{X}_{min}$ and $\hat{X}_{max}$ then follow a pairwise PVO-based PEE for embedding.

Once $\hat{X}_{min}$ and $\hat{X}_{max}$ are obtained, we pairwise expand pixels of each set in the backward embedding. In other words, both $\hat{X}_{min}$ and $\hat{X}_{max}$ are individually pairwise partitioned, sorted and used for embedding. For example, a pixel-pair $[\hat{l}_1, \hat{l}_2] \in \hat{X}_{min}$ with sorting becomes $[\hat{l}_{\sigma(1)}, \hat{l}_{\sigma(2)}]$. These partitioning and sorting also apply to $\hat{X}_{max}$, and thus, for each pixel-pair $[\hat{h}_1, \hat{h}_2] \in \hat{X}_{max}$, $[\hat{h}_{\sigma(1)}, \hat{h}_{\sigma(2)}] \leftarrow \sigma(\hat{h}_1, \hat{h}_2)$. We then predict $\hat{l}_{\sigma(2)}$ from $\hat{l}_{\sigma(1)}$ for $\hat{X}_{min}$ using Eq. (7a)–(7c), and predict $\hat{h}_{\sigma(1)}$ from $\hat{h}_{\sigma(2)}$ for $\hat{X}_{max}$ using Eq. (8a)–(8c). This prediction will increase the value of predict $\hat{h}_{\sigma(1)}$ from $\hat{h}_{\sigma(2)}$ for $\hat{X}_{max}$ using Eq. (8a)–(8c). This prediction will increase the value of

**Algorithm 1** $dPVO \cdot encode(\cdot)$

**Require:** $\hat{I}$
**Ensure:** $\hat{X}_{max}, \hat{X}_{min}, LM$
1: $\hat{X}_{min} \leftarrow empty$
2: $\hat{X}_{max} \leftarrow empty$
3: $LM \leftarrow zeros(2, k)$
4: $(M, N) \leftarrow size\ \hat{I}$
5: **for** all $k = 1$ to $\frac{M \times N}{3}$ **do**
6: $\quad [\hat{x}^k_{\sigma(1)}, \hat{x}^k_{\sigma(2)}, \hat{x}^k_{\sigma(3)}] \leftarrow sort\ [\hat{x}^k_1, \hat{x}^k_2, \hat{x}^k_3] \in \hat{I}$
7: $\quad$ **if** $\hat{x}^k_{\sigma(2)} - \hat{x}^k_{\sigma(1)} > 1$ **then**
8: $\quad\quad \hat{X}_{min} \leftarrow append(\hat{X}_{min}, \hat{x}^k_{\sigma(1)})$
9: $\quad$ **else**
10: $\quad\quad LM(1, k) \leftarrow 1$
11: $\quad$ **end if**
12: $\quad$ **if** $\hat{x}^k_{\sigma(3)} - \hat{x}^k_{\sigma(2)} > 1$ **then**
13: $\quad\quad \hat{X}_{max} \leftarrow append(\hat{X}_{max}, \hat{x}^k_{\sigma(3)})$
14: $\quad$ **else**
15: $\quad\quad LM(2, k) \leftarrow 1$
16: $\quad$ **end if**
17: **end for**
18: **return** $\hat{X}_{max}, \hat{X}_{min}, LM$

---

$\hat{l}_{\sigma(2)} \in \hat{X}_{min}$ by 0 or 1. Since all the pixel values in $\hat{X}_{min}$ have already decreased in the forward embedding by the value of 0 or 1, the backward embedding thus can partially restore the effect of the forward embedding resulting in lower distortion in the embedded image. Contrariwise, for applying the backward embedding to the pixel-pairs in $\hat{X}_{max}$, the lower pixel value $\hat{h}_{\sigma(1)}$ is predicted from $\hat{h}_{\sigma(2)}$. Thereby, we compute the set of expanded pixels, $\{\hat{\hat{l}}_{\sigma(2)}\}$ for $\hat{X}_{min}$ and $\{\hat{\hat{h}}_{\sigma(1)}\}$ for $\hat{X}_{max}$ to generate the expanded minimum-and maximum groups, $\hat{\hat{X}}_{min}$ and $\hat{\hat{X}}_{max}$, respectively. The final embedded image, $\hat{\hat{I}}$ is obtained by updating $\hat{I}$ with $\hat{\hat{X}}_{min}$ and $\hat{\hat{X}}_{max}$.

$$e_{xmin} = \hat{l}_{\sigma(2)} - \hat{l}_{\sigma(1)} \text{ for all } (\hat{l}_{\sigma(1)}, \hat{l}_{\sigma(2)}) \in \hat{X}_{min} \tag{7a}$$

$$\hat{e}_{xmin} = \begin{cases} e_{xmin}, & \text{if } e_{xmin} = 0 \\ e_{xmin} + b, & \text{if } e_{xmin} = 1 \\ e_{xmin} + 1, & \text{if } e_{xmin} > 1 \end{cases} \tag{7b}$$

$$\hat{\hat{l}}_{\sigma(2)} = \hat{l}_{\sigma(1)} + \hat{e}_{xmin} \tag{7c}$$

$$e_{xmax} = \hat{h}_{\sigma(1)} - \hat{h}_{\sigma(2)} \text{ for all } (\hat{h}_{\sigma(1)}, \hat{h}_{\sigma(2)}) \in \hat{X}_{max} \tag{8a}$$

$$\hat{e}_{xmax} = \begin{cases} e_{xmax}, & \text{if } e_{xmax} = 0 \\ e_{xmax} - b, & \text{if } e_{xmax} = -1 \\ e_{xmax} - 1, & \text{if } e_{xmax} > -1 \end{cases} \tag{8b}$$

$$\hat{\hat{h}}_{\sigma(1)} = \hat{h}_{\sigma(2)} + \hat{e}_{xmax} \tag{8c}$$

## Data extraction and image recovery

Data extraction and image recovery are inverse of embedding of our proposed RDH scheme. This means, data is first extracted with the inverse of backward embedding followed by the inverse of forward embedding. The input image to the decoder is partitioned into a non-overlapping image-block of size $1 \times 3$, and each block's pixels are sorted in either ascending or descending order. From the reserved pixels, the location map, $LM$ is extracted. From the sorted pixels of each block and using $LM$, the sets of maximum and minimum expanded pixels, $\hat{X}_{max}$ and $\hat{X}_{min}$, respectively are determined using Algorithm 2.

---

**Algorithm 2** $dPVO \cdot decode(\cdot)$

---

**Require:** $\hat{I}$ **Ensure:** $\hat{X}_{max}, \hat{X}_{min}$
1: $\hat{X}_{min} \leftarrow empty$
2: $\hat{X}_{max} \leftarrow empty$
3: $LM \leftarrow LMext(\hat{I})$
4: $(M, N) \leftarrow size(\hat{I})$
5: **for** all $k = 1$ to $\frac{M \times N}{3}$ **do**
6: $\quad [\hat{x}^k_{\sigma(1)}, \hat{x}^k_{\sigma(2)}, \hat{x}^k_{\sigma(3)}] \leftarrow sort([\hat{x}^k_1, \hat{x}^k_2, \hat{x}^k_3] \in \hat{I})$
7: $\quad$ **if** $\hat{x}^k_{\sigma(2)} - \hat{x}^k_{\sigma(1)} > 1$ & $LM(1, k) = 1$ **then**
8: $\quad\quad \hat{X}_{min} \leftarrow append(\hat{X}_{min}, \hat{x}^k_{\sigma(1)})$
9: $\quad$ **end if**
10: $\quad$ **if** $\hat{x}^k_{\sigma(3)} - \hat{x}^k_{\sigma(2)} > 1$ & $LM(2, k) = 1$ **then**
11: $\quad\quad \hat{X}_{max} \leftarrow append(\hat{X}_{max}, \hat{x}^k_{\sigma(3)})$
12: $\quad$ **end if**
13: **end for**
14: **return** $\hat{X}_{max}, \hat{X}_{min}$

---

Extraction of the embedded data bits, and recovery of $\hat{X}_{min}$ and $\hat{X}_{max}$ from $\hat{\hat{X}}_{min}$ and $\hat{\hat{X}}_{max}$, respectively are carried out using Eq. (9) and Eq. (10). We start with computing the errors from each pixel-pair in $\hat{\hat{X}}_{min}$ and $\hat{\hat{X}}_{max}$ using Eq. (9a) and Eq. (9b), respectively. Embedded bits are extracted using the error-values and conditions in Eq. (9c). The higher pixel, $\hat{\hat{l}}_{\sigma(2)}$ of each pixel-pair in $\hat{\hat{X}}_{min}$ are restored to $\hat{l}_{\sigma(2)}$ using Eq. (10a). Similarly, $\hat{h}_{\sigma(1)} \in \hat{X}_{max}$ is restored from $\hat{\hat{h}}_{\sigma(1)} \in \hat{\hat{X}}_{max}$ using Eq. (10b). Thereby, we can restore $\hat{X}_{max}$ and $\hat{X}_{min}$ from $\hat{\hat{X}}_{max}$ and $\hat{\hat{X}}_{min}$, respectively to finally compute $\hat{I}$ from $\hat{\hat{I}}$.

$$\hat{\hat{e}}_{xmin} = \hat{\hat{l}}_{\sigma(2)} - \hat{\hat{l}}_{\sigma(1)} \text{ for all } (\hat{\hat{l}}_{\sigma(1)}, \hat{\hat{l}}_{\sigma(2)}) \in \hat{\hat{X}}_{min} \quad (9a)$$

$$\hat{\hat{e}}_{xmax} = \hat{\hat{h}}_{\sigma(1)} - \hat{\hat{h}}_{\sigma(2)} \text{ for all } (\hat{\hat{h}}_{\sigma(1)}, \hat{\hat{h}}_{\sigma(2)}) \in \hat{\hat{X}}_{max} \quad (9b)$$

$$b = \begin{array}{ll} \hat{\hat{e}}_{xmin} - 1, & \text{if } 1 \leq \hat{\hat{e}}_{xmin} \leq 2 \\ -\hat{\hat{e}}_{xmax} - 1, & \text{if } -2 \leq \hat{\hat{e}}_{xmax} \leq -1 \end{array} \quad (9c)$$

For all $\hat{l}_{\sigma(2)} \in \hat{X}_{min}$ & $\hat{\hat{l}}_{\sigma(2)} \in \hat{\hat{X}}_{min}$ :

$$\hat{l}_{\sigma(2)} = \begin{cases} \hat{\hat{l}}_{\sigma(2)}, & \text{if } \hat{\hat{e}}_{xmin} = 0 \\ \hat{\hat{l}}_{\sigma(2)} - b, & \text{if } 1 \leq \hat{\hat{e}}_{min} \leq 2 \\ \hat{\hat{l}}_{\sigma(2)} - 1, & \text{if } \hat{\hat{e}}_{min} > 2 \end{cases} \quad (10a)$$

For all $\hat{h}_{\sigma(1)} \in \hat{X}_{max}$ & $\hat{\hat{h}}_{\sigma(1)} \in \hat{\hat{X}}_{max}$ :

$$\hat{h}_{\sigma(1)} = \begin{cases} \hat{\hat{h}}_{\sigma(1)}, & \text{if } \hat{\hat{e}}_{xmax} = 0 \\ \hat{\hat{h}}_{\sigma(1)} + b, & \text{if } -2 \leq \hat{\hat{e}}_{max} \leq -1 \\ \hat{\hat{h}}_{\sigma(1)} + 1, & \text{if } \hat{\hat{e}}_{max} < -2 \end{cases} \quad (10b)$$

The original image is finally restored with the inverse of our first phase embedding, which follows the extraction principle of Jung's scheme (see Sec. Jung's minimum PVO-based RDH scheme). This extraction phase starts with partitioning $\hat{I}$ into non-overlapping image-blocks. For each block, its pixels $[\hat{x}_1, \hat{x}_2, \hat{x}_3]$ are then sorted to $[\hat{x}_{\sigma(1)}, \hat{x}_{\sigma(2)}, \hat{x}_{\sigma(3)}]$. Respective errors, $\hat{e}_{max}$ and $\hat{e}_{min}$ are computed using Eq. (11) followed by the data-bit extraction and pixel recovery using Eq. (5b) to Eq. (5d). Upon the extraction of all data-bits, they are marged to the data-bits extracted in the earlier phase. Similarly, the original image $I$ is obtained by updating $\hat{I}$ with the restored values of $\hat{x}_{\sigma(3)}$ and $\hat{x}_{\sigma(1)}$ of each image-block.

$$\hat{e}_{max} = \hat{x}_{\sigma(3)} - \hat{x}_{\sigma(2)} \quad (11a)$$
$$\hat{e}_{min} = \hat{x}_{\sigma(1)} - \hat{x}_{\sigma(2)} \quad (11b)$$

## An working example

We now illustrate the step-by-step processing of the proposed dPVO scheme with an simple example. The embedding and decoding processes for both the *forward* and *backward* phases are presented in Fig 2 and Fig 4, respectively. Thereby we demonstrate the restoration potential of the *backward* embedding in Fig 3.

The example in Fig 2 operates on a tiny input image consisting a set of $(1 \times 3)$-sized blocks that we call *input blocks* in the figure, which transforms to the *sorted blocks* after block-wise sorting using $\sigma(.)$. For instance, in Fig 2, the first *input block*, [161, 160, 162] becomes [160, 161, 162] as the *sorted block*, which is then processed for the expansion (*i.e.*, *embedding* or *unit-value shifting*) in both the *forward* and *backward* embedding. In *forward* phase, for the sorted bolck [160, 161, 162], we get the prediction errors, $e_{min}$ = -1 and $e_{max}$= 1 using Eq. (4a) and Eq. (4b). Based on the values of $e_{min}$ and $e_{max}$, the expanded versions of the prediction errors, $\hat{e}_{min}$ and $\hat{e}_{max}$ are obtained either embedding a data-bit, b or by unit shifting of the error-values using Eq. (4c) and Eq. (4d). Let us consider a data as an example "11101....." to be embedded. Using Eq. (4c) and Eq. (4d), we get $\hat{e}_{min}$ = -2 and $\hat{e}_{max}$ = 2 (embedding the first two bits of the data, "11"). Thus the sorted block after forward embedding is obtained as [159, 161, 163]. Similarly, the second block *input block*, [159, 161, 166], after sorting and forward embedding becomes [158, 160, 161], embedding two data-bits "10" as shown in the figure. Unlike the first and second block, no data is embedded in the third *input block*, [201, 205, 242]. Rather unit expansion as per Eq. (4c) and Eq. (4d) is done due to the values of $e_{min}$ and $e_{max}$ being "-4" and "37", respectively. Thus corresponding the embedded block becomes [200, 205, 243]. Similarly, embedding one bit data in the maximum pixel and shifting the minimum pixel by unity, the forth *input block*, [242, 243, 199] becomes [198, 242, 244].

The backward embedding operates on the forwarded embedded blocks with an aim to restoring the changes in forward phase. According to algorithm 1, from the forward embedded blocks, the minimum and maximum set of pixels, $\hat{X}_{min}$ and $\hat{X}_{max}$ are obtained based on the expanded prediction errors $\hat{e}_{min}$ and $\hat{e}_{max}$, respectively. We get the minimum set $\hat{X}_{min}$ = [159, 158, 200, 198] and $\hat{X}_{max}$ = [163, _, 243, 244]. Now, $\hat{X}_{min}$ and $\hat{X}_{max}$ are pairwise partitioned, sorted and finally used for embedding. For example, the first sorted pair of the minimum pixels set, $[\hat{l}_{\sigma(1)}, \hat{l}_{\sigma(2)}]$ = [158, 159] and corresponding $\hat{e}_{xmin}$= 1, thus one bit of data will be embedded and $\hat{\hat{e}}_{xmin}$ will be 2 (as the data-bit is '1'), and finally $\hat{l}_{\sigma(2)}$ will be 160 after embedding. In the similar fashion, all the pairs in the minimum and maximum sets are considered, sorted and after checking the value of corresponding $\hat{e}_{xmin}$ or $\hat{e}_{xmax}$, new value of $\hat{\hat{e}}_{xmin}$ or $\hat{\hat{e}}_{xmax}$ are calculated and embedded minimum and maximum sets, $\hat{\hat{X}}_{min}$ and $\hat{\hat{X}}_{max}$ are determined. Finally, de-sorting the pixels block-wise, the final embedded image is obtained.

The ability of the backward embedding for restoring the changes occurred in the forward phase is demonstrated in Fig 3. In this example, it is clearly seen from Fig 3 (a), *i.e.*, $I - \hat{I}$ that seven pixel values have been changed in the forward phase. From Fig 3 (b), *i.e.*, in $I - \hat{\hat{I}}$ the restored pixels are marked as blue and green cells. Our backward embedding could restore three pixels out of seven forward embedded pixels, which will surely increase the image quality with slightly higher embedding capacity.

The decoding process of the proposed scheme is illustrated in Fig 4 with the same tiny image block. The decoding process is exactly reverse to encoding. So, it starts with sorting the three pixels in each block. Using the sorted embedded blocks and the location map, the expanded minimum and maximum sets, $\hat{\hat{X}}_{min}$ and $\hat{\hat{X}}_{max}$ are determined. Then we find $\hat{e}_{xmin}$ or $\hat{e}_{xmax}$ from each consecutive pair of $\hat{X}_{min}$ and $\hat{X}_{max}$. The value of $\hat{e}_{xmin}$ or $\hat{e}_{xmax}$ indicates whether any data-bit is embedded or not as per Eq. (9c). The value of $\hat{l}_{\sigma(2)}$ is calculated from $\hat{\hat{l}}_{\sigma(2)}$ using Eq. (10a) and $\hat{h}_{\sigma(1)}$ is obtained from , $\hat{\hat{h}}_{\sigma(1)}$ using Eq. (10b). Thus from the sorted minimum embedded pixels group, [160, 158, 201, 198], we obtain the forward embedded column [159, 158, 200, 198]. Similarly, from the sorted embedded maximum pixel group, [162, 161, 243, 244],

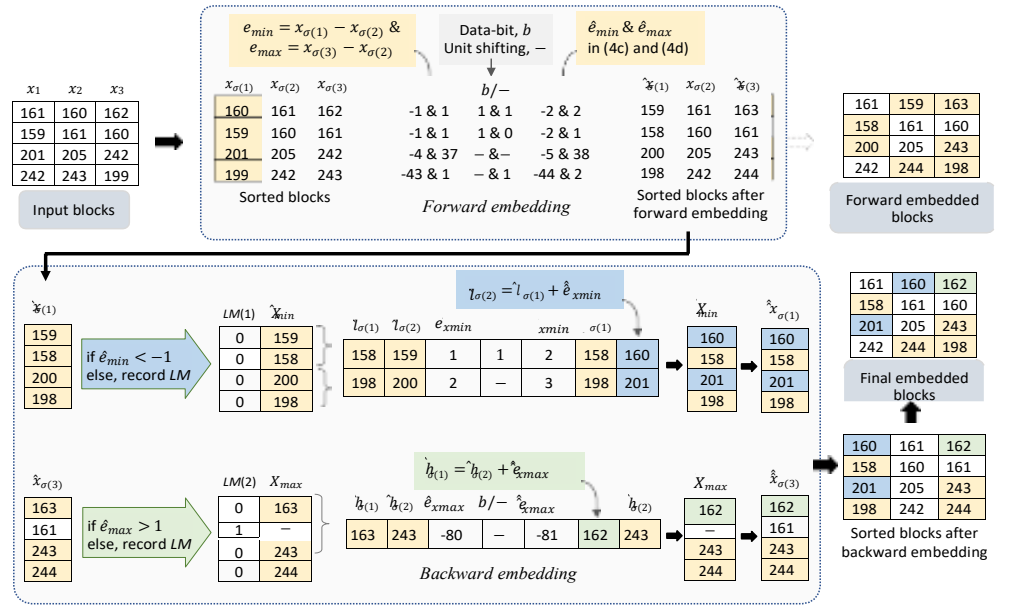

**Fig 2. Example of the *forward* and *backward* embedding (the *yellow-cells* represent the pixels and their changes in *forward* embedding, and the *blue-* and *green-cells* represent the pixels and their changes in *backward* embedding)**

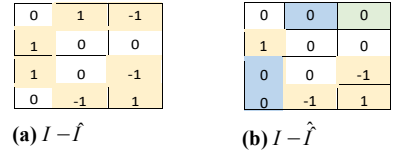

**(a)** $I - \hat{I}$  **(b)** $I - \hat{I}$

**Fig 3.** Restoration ability of *backward* embedding: the changes in pixel-values after (a) *forward* embedding (*yellow cells*) and (b) *backward* embedding (*blue & green cells*)

the forward embedded column [163, 161, 243, 244] is obtained. Combining these columns we obtain the backward recovered blocks which is identical to the forward embedded blocks. These recovered blocks are then utilized to determine the value of $e_{min}$ and $e_{max}$, which in turn helps restore the original sorted blocks using Eq. (5b) to Eq. (5d) and to extract the embedded data-bits using Eq. (5a). Finally a de-sorting process yields the original input image.

# Experimental results and analysis

In this section, we present the performance of our RDH scheme for its analysis and validation. We used the popular test-images of size $512 \times 512 \times 8$ from the USC-SIPI [57] for this performance evaluation. All the other illustrations (figures and plots) are also available on Figshare [58]. We determine both the embedding-capacity and embedding-rate in terms of total embedded bits and bit per pixels (bpp), respectively. For embedding, a set of pseudo-random bits is generated as *data*. Implementations are carried out using MATLAB R2016b with a 1.3 GHz Intel Core i5 CPU, 4 GB memory.

We have evaluated the embedded image quality in terms of a popular objective visual quality metric, peak signal to noise ratio (PSNR) defined in Eq. (12). Here, $M \times N$ is the image size, and $I(i, j)$ and $I^t(i, j)$ are the pixel values of position $(i, j)$ in an original image and its embedded version, respectively. $L$ is the dynamic range of the pixel values.

$$\text{MSE} = \frac{\sum_{j=1}^{N} \sum_{i=1}^{M} \left(I^t(i, j) - I(i, j)\right)^2}{MN} \tag{12a}$$

$$\text{PSNR} = 10 \log \frac{L^2}{\text{MSE}} \tag{12b}$$

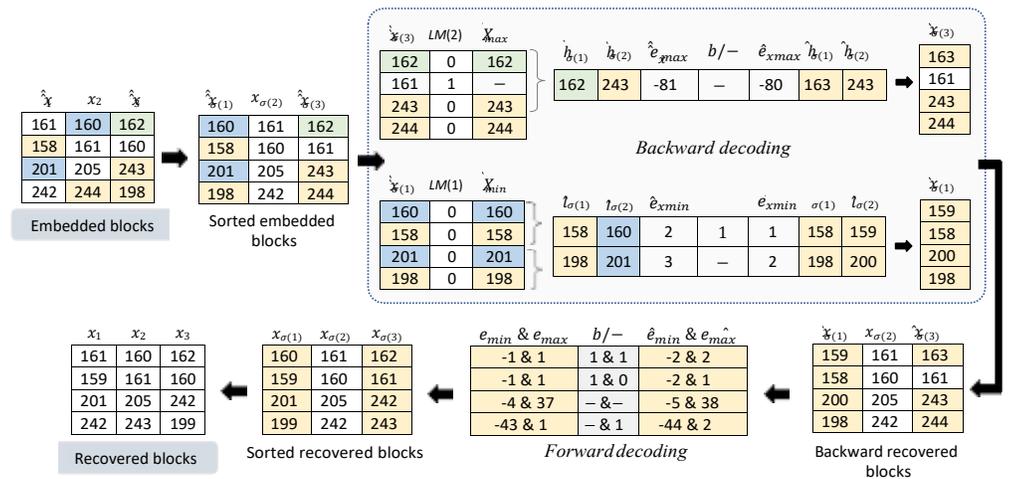

**Fig 4.** Example of the *forward* and *backward* decoding

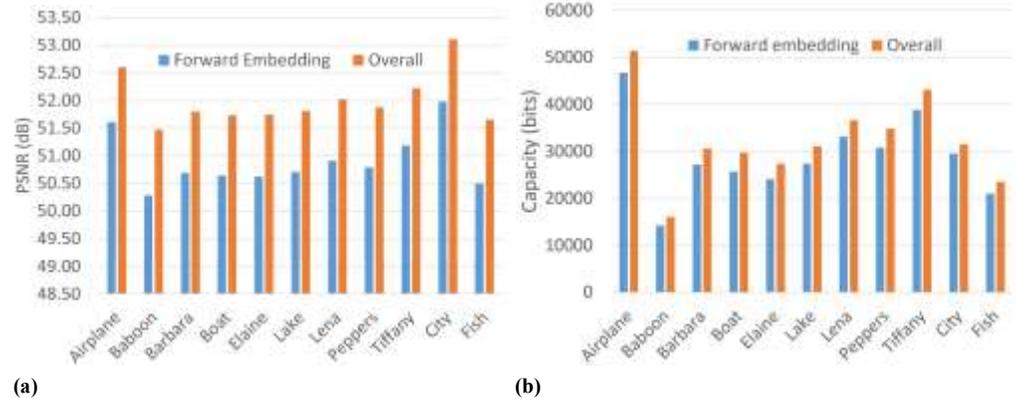

(a)                                                            (b)

**Fig 5. Rate-distortion performance with proposed dPVO for test images: (a) PSNR and (b) Capacity**

It is evident form Table 1, with the basic PVO, *i.e.*, for forward embedding, an average PSNR of 50.90 dB is obtained. With the application of backward embedding, the embedded image is slightly restored to mitigate the distortion occurred in the forward phase. Consequently, the overall average PSNR is obtained as 52 dB with a significant improvement of 2.16%. Moreover, the backward embedding makes more room for data, thus the overall embedding capacity is increased as depicted from Table 1. The inclusion of backward embedding offers 11.98% increment in the embedding capacity in average for the test images.

The merit of backward embedding in improving the image quality and embedding capacity is further demonstrated graphically in Fig 5 for the test images. The improvement in PSNR and capacity is depicted with the inclusion of the backward phase in the embedding process. Example of embedded images and their decoded versions are illustrated in Fig 6. The decoded image, by definition of our decoding principle, should be identical to the input image, which is verified for all the test images and can be roughly observed with the given examples.

Overall embedding rate-distortion performance of our scheme is evaluated and presented in Table 1. Average PSNR (dB) value of 50.90 with embedding capacity of 32.33 kbit is obtained

**Table 1. Improvement in image quality and capacity for the test images**

| Image | PSNR (dB) | | | Capacity (bits) | | |
|---|---|---|---|---|---|---|
| | Forward phase | Overall | Improvement (%) | Forward phase | Overall | Improvement (%) |
| Airplane | 51.61 | 52.60 | 1.93% | 46741 | 51376 | 9.92% |
| Baboon | 50.29 | 51.47 | 2.35% | 14203 | 16011 | 12.73% |
| Barbara | 50.69 | 51.80 | 2.19% | 27162 | 30559 | 12.51% |
| Boat | 50.64 | 51.73 | 2.15% | 25635 | 29686 | 15.80% |
| Elaine | 50.62 | 51.74 | 2.21% | 24081 | 27367 | 13.65% |
| Lake | 50.71 | 51.81 | 2.17% | 27387 | 31007 | 13.22% |
| Lena | 50.91 | 52.02 | 2.18% | 33111 | 36607 | 10.56% |
| Peppers | 50.79 | 51.88 | 2.13% | 30758 | 34797 | 13.13% |
| Tiffany | 51.18 | 52.22 | 2.03% | 38813 | 43158 | 11.19% |
| City | 51.98 | 53.11 | 2.17% | 29513 | 31528 | 6.83% |
| Fish | 50.50 | 51.65 | 2.28% | 20969 | 23534 | 12.23% |
| **Average** | **50.90** | **52.00** | **2.16%** | **28943** | **32330** | **11.98%** |

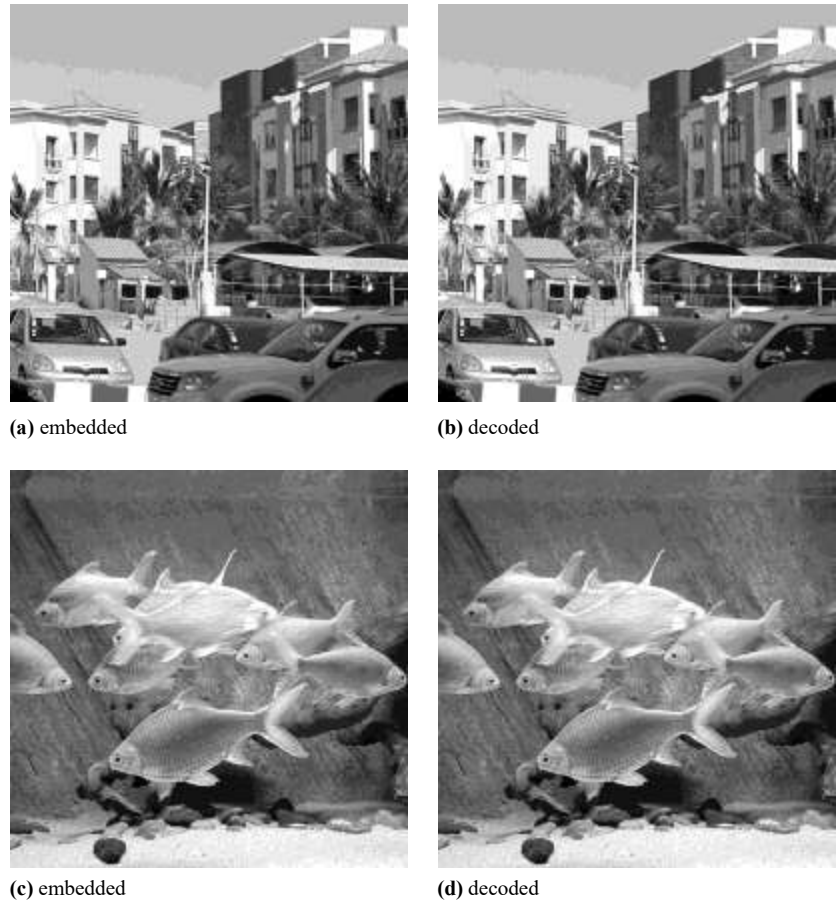

**(a)** embedded      **(b)** decoded

**(c)** embedded      **(d)** decoded

**Fig 6. Example of embedded and decoded versions of test images of size $512 \times 512 \times 8$, (a,b)** *City* **and (c,d)** *Fish* **(Original test images are from author's collection)**

for the test-images. Table 2 represents a statistical count of pixels undergoing different operations in forward and backward phases, like shifting, expansion, embedding or remaining unchanged. From Table 2, it is observed that only 46% pixels in the forward embedded image remain unchanged in average with reference to the input image in terms of pixel intensity value. Backward embedding ensures 88% pixels unchanged in average with respect to the forward embedded image, while the rest 12% pixels are backward shifted or embedded with data-bits. This backward phase yields the restoration of a significant amount of pixels providing overall 58% pixels unchanged in average with respect to the input image. Therefore, the inclusion of backward embedding significantly helps to improve the image quality.

    We compare the performance of our scheme with a number of popular and recent RDH schemes: He *et al*. (2021) [54], Kumar & Jung (2020) [53], He *et al*. (2018) [38], Jung (2017) [36], Ou *et al*. (2016) [34], Wang *et al*. (2015) [33], Qu & Kim (2015) [32], Peng *et al*. (2014) [30], Li *et al*. (2013) [29] and Sachnev *et al*. (2009) [19]. Embedded image quality for embedding 10 kbit and 20 kbit of data in the test-images is evaluated and compared with the relevant PVO and PEE based RDH schemes in Table 3 and Table 4, respectively. Generally, in contrast to all those schemes, in both cases of embedding 10 kbit and 20 kbit data, PSNR values of ours remain higher. A better average value of PSNR for our scheme is always obtained than that of the Jung's scheme, which our scheme is directly built on. For example, an average PSNR value of our scheme remains about 7% higher than the Jung's scheme in case of 10 kbit of

Table 2. Statistics of the embedded, unchanged, expanded and restored pixels with the test images.

| Images | Forward embedding | | | | | | | | | Backward embedding | | | | | | | | | | Overall | | | |
|---|---|---|---|---|---|---|---|---|---|---|---|---|---|---|---|---|---|---|---|---|---|---|---|
| | Unchanged pixels | | Expanded pixels | | | | | Bits embedded | | Skipped pixels | | Unchanged pixels | | Expanded pixels | | | | | Bits Embedded | | Unchanged pixels | | Expanded Pixels | |
| | | | Embedded | | Shifted | | | | | | | | | Embedded | | Shifted | | | | | | | | |
| | Count | Rate | Count "0" | Count "1" | Count | Rate | | Count | Rate | Count | Rate | Count | Rate | Count "0" | Count "1" | Count | Rate | Count | Rate | Count | Rate | Count | Rate |
| Airfield | 120870 | 46% | 14953 | 14577 | 126697 | 48% | | 29530 | 11% | 33830 | 13% | 228923 | 87% | 1262 | 1257 | 30940 | 12% | 2519 | 1% | 153067 | 58% | 109077 | 42% |
| Airplane | 143351 | 55% | 23537 | 23204 | 95589 | 36% | | 46741 | 18% | 56311 | 21% | 237010 | 90% | 2340 | 2295 | 21815 | 8% | 4635 | 2% | 167461 | 64% | 94683 | 36% |
| Baboon | 101520 | 39% | 7260 | 6943 | 153681 | 59% | | 14203 | 5% | 14480 | 6% | 223053 | 85% | 907 | 901 | 37166 | 14% | 1808 | 1% | 139587 | 53% | 122557 | 47% |
| Barbara | 115676 | 44% | 13764 | 13398 | 133070 | 51% | | 27162 | 10% | 28636 | 11% | 228380 | 87% | 1702 | 1695 | 31045 | 12% | 3397 | 1% | 148416 | 57% | 113728 | 43% |
| Boat | 114147 | 44% | 13012 | 12623 | 135374 | 52% | | 25635 | 10% | 27107 | 10% | 228495 | 87% | 2016 | 2035 | 30590 | 12% | 4051 | 2% | 146772 | 56% | 115372 | 44% |
| Elaine | 113379 | 43% | 12241 | 11840 | 136925 | 52% | | 24081 | 9% | 26339 | 10% | 227523 | 87% | 1629 | 1657 | 31940 | 12% | 3286 | 1% | 146976 | 56% | 115168 | 44% |
| Lake | 116376 | 44% | 13881 | 13506 | 132262 | 50% | | 27387 | 10% | 29336 | 11% | 228717 | 87% | 1800 | 1820 | 30583 | 12% | 3620 | 1% | 148779 | 57% | 113365 | 43% |
| Lena | 122894 | 47% | 16729 | 16382 | 122868 | 47% | | 33111 | 13% | 35854 | 14% | 229896 | 88% | 1791 | 1705 | 29519 | 11% | 3496 | 1% | 154118 | 59% | 108026 | 41% |
| Peppers | 119164 | 45% | 15559 | 15199 | 127781 | 49% | | 30758 | 12% | 32124 | 12% | 229758 | 88% | 2019 | 2020 | 29342 | 11% | 4039 | 2% | 150526 | 57% | 111618 | 43% |
| Tiffany | 131275 | 50% | 19637 | 19176 | 111693 | 43% | | 38813 | 15% | 44235 | 17% | 233507 | 89% | 2114 | 2231 | 25382 | 10% | 4345 | 2% | 158888 | 61% | 103256 | 39% |
| Zelda | 122146 | 47% | 16986 | 16658 | 123340 | 47% | | 33644 | 13% | 35106 | 13% | 229724 | 88% | 1710 | 1621 | 29775 | 11% | 3331 | 1% | 153542 | 59% | 108602 | 41% |
| City | 153284 | 58% | 14945 | 14568 | 94292 | 36% | | 29513 | 11% | 66244 | 25% | 236556 | 90% | 997 | 1018 | 23546 | 9% | 2015 | 1% | 177848 | 68% | 84296 | 32% |
| Fish | 109294 | 42% | 10693 | 10276 | 142574 | 54% | | 20969 | 8% | 22254 | 8% | 225842 | 86% | 1269 | 1296 | 33982 | 13% | 2565 | 1% | 144572 | 55% | 117572 | 45% |
| **Average** | **121798** | **46%** | **14861** | **14488** | **125857** | **48%** | | **29350** | **11%** | **34758** | **13%** | **229799** | **88%** | **1658** | **1658** | **29664** | **11%** | **3316** | **1%** | **153119** | **58%** | **109025** | **42%** |

embedding for USC-SIPI image sets as shown in Table 3. Similarly, for embedding 20 kbit of data in different test images, this improvement in the average rate-distortion performance is also evident in Table 4.

The percentage improvement in image quality offered by our scheme for embedding both 10 kbit and 20 kbit data is also graphically compared with other schemes in Fig 7. For example, for 10 kbit data embedding, our scheme offers 4.3%, 1.6%, 0.8%, 0.3%, 0.3%, 0.7%, 7%, 0.3%, 2.8% and -0.9% higher PSNR value than Sachnev *et al*. [19], Li *et al*. [29], Peng *et al*. [30], Qu & Kim [32], Wang *et al*. [33], Ou *et al*. [34], Jung [36], He *et al*. [38], Kumar & Jung (2020) [53] and He *et al*. (2021) [54], respectively. With a similar trend, 4.5%, 2.5%, 1.9%, 1.1%, 1.2%, 1.7%, 6.7%, 1.2%, 2.2% and 1.2% better PSNR is obtained with our scheme for embedding 20 kbit data compared to the schemes in [19], [29], [30], [32], [33], [34], [36], and [38], Kumar & Jung (2020) [53] and He *et al*. (2021) [54], respectively. It is observed that in most of the cases, improvement rate for embedding 20 kbit data is higher than that for embedding 10 kbit. This means, ours scheme has a tendency to improve image quality for a higher embedding rate.

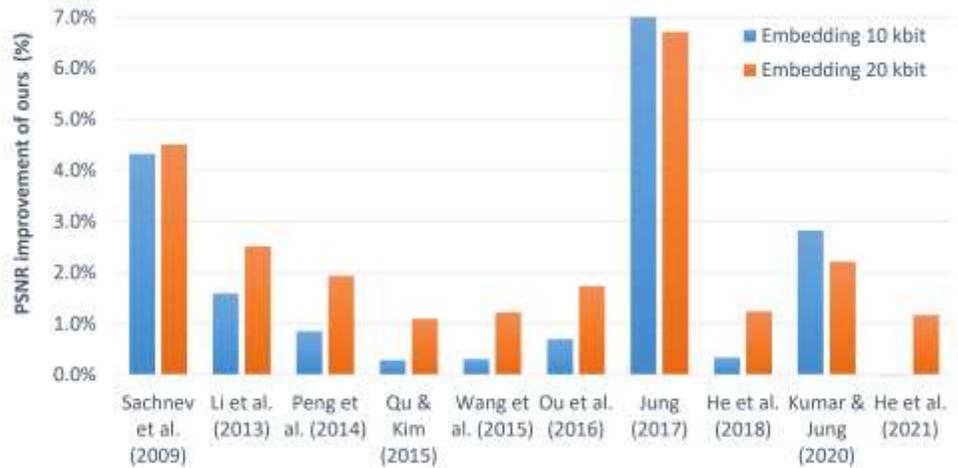

**Fig 7. Average improvement for embedding 10 kbit and 20 kbit over the schemes: He *et al*. (2021) [54], Kumar & Jung (2020) [53], He *et al*. (2018) [38], Jung (2017) [36], Ou *et al*. (2016) [34], Wang *et al*. (2015) [33], Qu & Kim (2015) [32], Peng *et al*. (2014) [30] and Sachnev *et al*. (2009) [19].**

Additionally, to visualize the trend of the overall performance of our scheme, embedding rate-distortion curve is compared with a few recent and popular RDH schemes [19, 30, 32–34, 36, 38, 53, 54] in Fig 8. We observe that our scheme has a trend to

**Table 3. PSNRs (dB) for embedding 10,000 bits in the USC-SIPI images.**

| Schemes | Sachnev *et al*. [19] | Li *et al*. [29] | Peng *et al*. [30] | Qu & Kim [32] | Wang *et al*. [33] | Ou *et al*. [34] | Jung [36] | He *et al*. [38] | Kumar & Jung [53] | He *et al*. [54] | Ours |
|---|---|---|---|---|---|---|---|---|---|---|---|
| Lena | 58.18 | 60.30 | 60.47 | 60.31 | 60.44 | 60.46 | 56.99 | 60.64 | 60.44 | 61.01 | 60.50 |
| Baboon | 54.15 | 53.52 | 53.55 | 54.21 | 54.50 | 54.16 | 51.62 | 54.00 | 54.95 | 59.92 | 55.19 |
| Barbara | 58.15 | 59.81 | 60.54 | 59.77 | 60.27 | 60.15 | 55.69 | 60.37 | 58.56 | 60.96 | 58.96 |
| Airplane | 60.37 | 62.00 | 62.96 | 63.68 | 63.41 | 63.14 | 58.63 | 63.45 | 60.86 | 59.92 | 63.65 |
| Peppers | 55.55 | 58.87 | 58.98 | 58.78 | 58.97 | 59.16 | 55.86 | 59.29 | 56.23 | 59.61 | 59.77 |
| Boat | 56.15 | 58.11 | 58.27 | 58.42 | 58.39 | 58.06 | 54.61 | 58.28 | 56.33 | 58.78 | 58.90 |
| Elaine | 56.12 | 56.81 | 57.36 | 58.72 | 58.10 | 57.36 | 54.65 | 57.67 | 56.65 | 58.78 | 58.56 |
| Lake | 56.65 | 58.21 | 58.87 | 59.76 | 59.50 | 59.23 | 55.75 | 59.71 | 57.95 | 60.43 | 59.51 |
| **Average** | **56.92** | **58.45** | **58.88** | **59.21** | **59.20** | **58.97** | **55.48** | **59.18** | **57.75** | **59.92** | **59.38** |

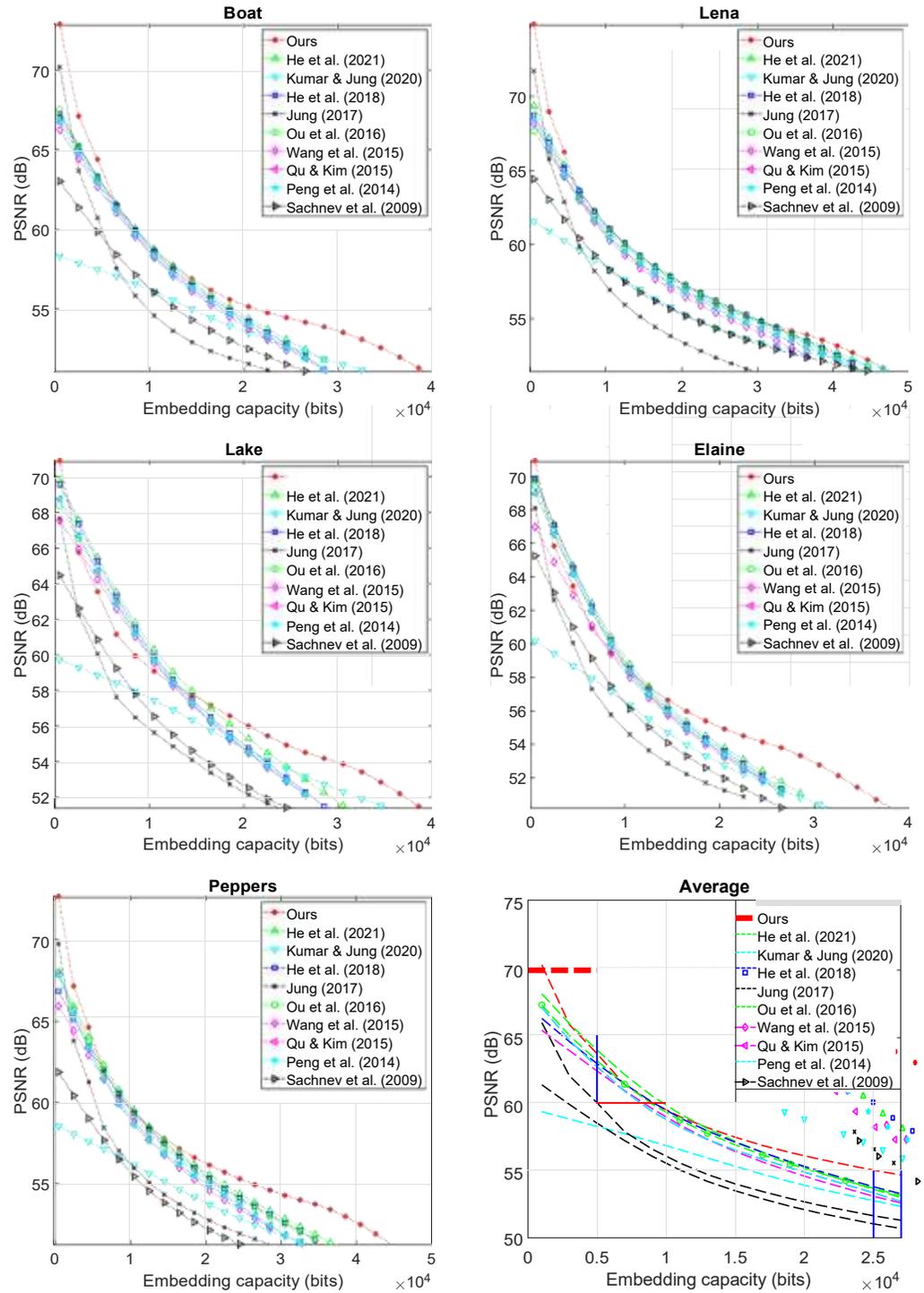

**Fig 8. Overall embedding rate-distortion performance comparison with the popular and recent RDH schemes of He *et al*. (2021) [54], Kumar & Jung (2020) [53], He *et al*. (2018) [38], Jung (2017) [36], Ou *et al*. (2016) [34], Wang et al. (2015) [33], Qu & Kim (2015) [32], Peng *et al*. (2014) [30] and Sachnev *et al*. (2009) [19].**

**Table 4. PSNRs (dB) for embedding 20,000 bits in the USC-SIPI images.**

| Schemes | Sachnev et al. [19] | Li et al. [29] | Peng et al. [30] | Qu & Kim [32] | Wang et al. [33] | Ou et al. [34] | Jung [36] | He et al. [38] | Kumar & Jung [53] | He et al. [54] | Ours |
|---|---|---|---|---|---|---|---|---|---|---|---|
| Lena | 55.03 | 56.21 | 56.54 | 56.67 | 56.65 | 56.6 | 53.40 | 56.81 | 55.22 | 57.34 | 56.92 |
| Baboon | – | – | – | – | – | – | – | – | 52.86 | – | 49.95 |
| Barbara | 55.04 | 54.69 | 56.2 | 55.63 | 56.50 | 55.89 | 51.99 | 56.09 | 55.86 | 57.09 | 55.75 |
| Airplane | 57.32 | 58.13 | 59.07 | 59.91 | 59.61 | 59.26 | 56.16 | 59.59 | 58.96 | 55.65 | 58.97 |
| Peppers | 52.30 | 54.72 | 54.77 | 54.98 | 54.81 | 54.93 | 52.67 | 55.1 | 53.96 | 55.67 | 56.23 |
| Boat | 52.65 | 53.34 | 53.84 | 54.21 | 53.96 | 53.72 | 51.69 | 54.07 | 53.76 | 54.51 | 55.11 |
| Elaine | 52.01 | 52.41 | 52.61 | 53.71 | 53.29 | 52.71 | 51.33 | 53.08 | 52.92 | 53.87 | 54.95 |
| Lake | 52.72 | 53.44 | 53.6 | 54.69 | 54.50 | 54.28 | 52.08 | 54.53 | 54.86 | 55.41 | 56.17 |
| **Avrage** | **53.87** | **54.92** | **55.23** | **55.69** | **55.62** | **55.34** | **52.76** | **55.61** | **55.08*** | **55.65** | **56.30*** |

*Average value is calculated without Baboon image.

improve embedded image quality over the higher embedding rate. This is because a higher rate of embedding requires a higher number of pixels to be embedded. Once the number of pixels becomes higher, the sizes of $X_{min}$ and $X_{max}$ in the second phase of embedding tend to be higher resulting in more restored pixels and better-embedded image quality. The average performance in Fig 8 also evidences the trend of improvement. We note that this trend of improvements in the rate-distortion performance of our RDH scheme discussed and illustrated above for a few test images also holds for the other test images we experimented with.

On the location map, despite embedding in two phases, the proposed scheme requires two location maps of total size of a single plane like a PEE based scheme. For example, the location map size for forward embedding in an image of size $512 \times 512$ is $\frac{512 \times 512}{3}$ bits to mark any image block of size $1 \times 3$ containing the boundary pixel(s). Besides, in the backward embedding, pair based-PEE is used requiring a location map of size $\frac{2 \times 512 \times 512}{3}$ bits. Thus, the total location map size is $512 \times 512$ bits, which is an implicit requirement of a typical PEE based RDH scheme as also seen in the schemes considered for the performance comparison. We also note that having the similar requirements of the location map, it holds similar effects on the embedding capacity of the concerned RDH schemes including the proposed one.

In summary, considering the overall rate-distortion performance, our proposed RDH scheme outperforms its baseline schemes: Jung (2017) [36], Peng et al. (2014) [30] and Li et al. (2013) [29]. Our scheme is based on the classic PVO and uses a simple scenario of image partitions of size $1 \times 3$, and thus we reasonably considered those schemes for baseline comparison. Moreover, a promising performance of our scheme is also observed, while we compared it to the other popular state-of-the-art PVO based RDH schemes [19,32–34,38,53,54].

## Conclusions

In this paper, we introduced a new RDH scheme with dPVO based backward embedding. With two phase-embedding, our RDH scheme first applies classic PVO based PEE on each non-overlapping image-block of size $1 \times 3$. The second phase embedding with dPVO based backward embedding is designed to partially restore the pixels predicted with expansion or embedding in the first phase. Our substantial experimental results demonstrated a promising performance of our proposed scheme and its improvement over the popular and state-of-the-art PVO-based RDH schemes. Particularly, our scheme demonstrated significantly better rate-distortion performance at the higher embedding rate compared to the state-of-the-art PVO-based RDH schemes.

Better embedded image quality at higher embedding rates means to have more potential for the applications that usually require high embedding capacity like electronic patient record hiding in medical images. In addition to the study of its specific application scenarios, future investigation on the proposed dPVO based principle of backward embedding may be worthwhile, for example, in the following areas: (*i*) its information theocratic analysis, (*ii*) optimization of its computational requirements for multi-level embedding, and (*iii*) developing its generalized framework for multi-level embedding and dynamic image-block partitioning.